\documentclass[runningheads]{llncs}
\usepackage{graphicx}

\usepackage{tikz}
\usepackage{comment}
\usepackage{amsmath,amssymb} 
\usepackage{color}
\usepackage{xcolor}
\usepackage{colortbl}
\usepackage{cite}
\usepackage{enumitem}
\usepackage{wrapfig}
\usepackage{floatflt}
\usepackage{booktabs}
\usepackage{amssymb}
\usepackage{pifont}
\newcommand{\xmark}{\ding{53}}

\def\zsw{\textcolor{black}}

\usepackage[accsupp]{axessibility}  

\usepackage[width=122mm,left=12mm,paperwidth=146mm,height=193mm,top=12mm,paperheight=217mm]{geometry}

\begin{document}
\pagestyle{headings}
\mainmatter
\def\ECCVSubNumber{710}  

\title{Open-world Semantic Segmentation for \\ LIDAR Point Clouds} 

%
\author{Jun Cen\inst{1} \and Peng Yun\inst{1} \and Shiwei Zhang$^{2*}$ \and Junhao Cai\inst{1} \and Di Luan\inst{1} \and \\ Michael Yu Wang\inst{1*} \and Ming Liu\inst{1} \and Mingqian Tang\inst{2}}
\authorrunning{J. Cen et al.}
%
\institute{The Hong Kong University of Science and Technology \and Alibaba Group \\
\email{\{jcenaa,pyun,jcaiaq,dluan\}@connect.ust.hk}
\email{\{mywang,eelium\}@ust.hk}\\
\email{\{zhangjin.zsw,mingqian.tmq\}@alibaba-inc.com}}
\maketitle

\let\thefootnote\relax\footnotetext{\scriptsize{$^*$Corresponding authors.}}
\let\thefootnote\relax\footnotetext{\scriptsize{Code is available at: \url{https://github.com/Jun-CEN/Open\_world\_3D\_semantic\_segmentation}}}

\begin{abstract}
\zsw{Current methods for} LIDAR semantic segmentation are not robust enough for real-world applications, \zsw{\textit{e.g.}}, autonomous driving, since it is {\it closed-set} and {\it static}.
The closed-set assumption makes the network only able to output labels of trained classes, even for objects never seen before, while a static network cannot update its knowledge base according to what it has seen. 
Therefore, in this work, we propose the {\it open-world semantic segmentation} task for LIDAR point clouds, which aims to 1) identify both old and novel classes using open-set semantic segmentation, 
and 2) gradually incorporate novel objects into the existing knowledge base using incremental learning without forgetting old classes. 
%
\zsw{For this purpose, we propose a \textbf{RE}dund\textbf{A}ncy c\textbf{L}assifier (REAL) framework to provide a general architecture for both the open-set semantic segmentation and incremental learning problems.} 
%
\zsw{The experimental results show that REAL can simultaneously achieves state-of-the-art performance in the open-set semantic segmentation task on the SemanticKITTI and nuScenes datasets, and alleviate the catastrophic forgetting problem with a large margin during incremental learning.
}

\end{abstract}

\section{Introduction}
\label{sec:intro}

\begin{figure}[t]
\centering
\includegraphics[width=0.99\textwidth]{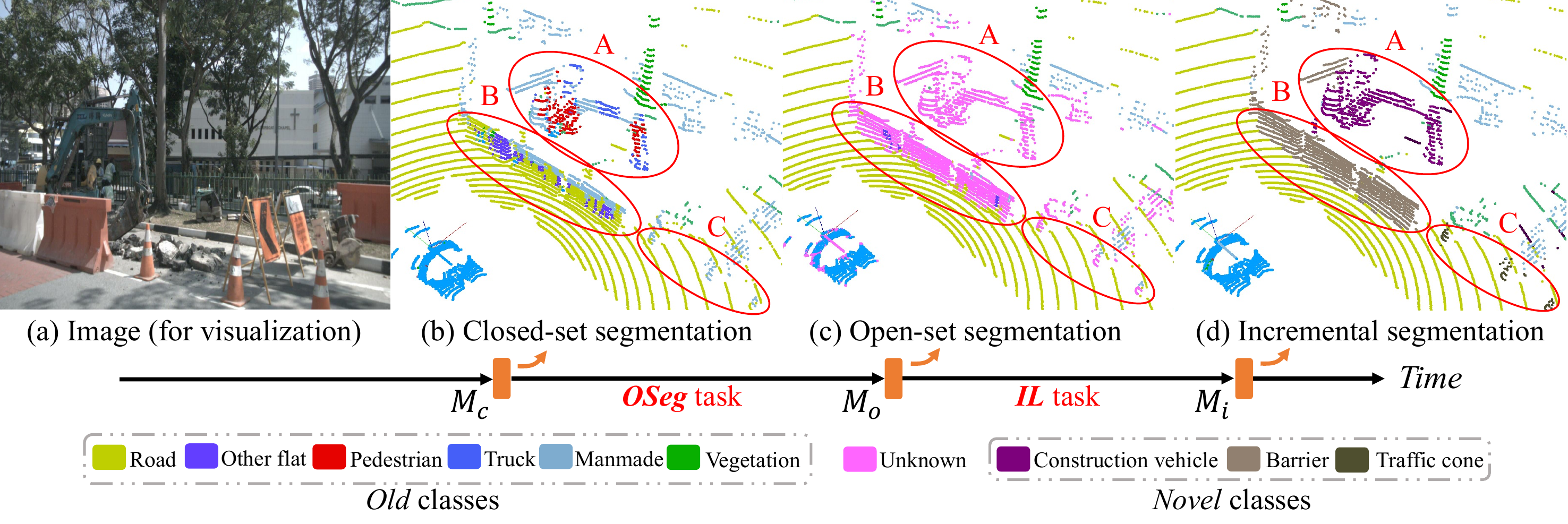}
\vspace{-0.3cm}
\caption{Closed-set model $\mathcal{M}_c$ wrongly assigns the labels of old classes to novel objects (A: construction vehicle is classified as the manmade, truck, and even pedestrian; B: barrier is classified as the road, manmade and other flat; C: traffic cone is classified as the manmade). After open-set semantic segmentation (OSeg) task, the open-set model $\mathcal{M}_o$ can identify the novel objects and assign the label \textit{unknown} for them. After incremental learning (IL) task, the model $\mathcal{M}_i$ can classify both old and novel classes.}
\label{fig:example_1}
\end{figure}

3D LIDAR sensors play an important role in the perception system of autonomous vehicles.
Semantic segmentation for LIDAR point clouds has grown very fast in recent years~\cite{cylindrical, 2-S3Net, polarnet, rangenet++}, benefiting from well-annotated datasets including SemanticKITTI~\cite{se1, se2, se3} and nuScenes~\cite{nuscenes}. 
However, existing methods for LIDAR semantic segmentation are all \textit{closed-set} and \textit{static}. 
The closed-set network regards all inputs as categories encountered during training, so it will assign the labels of old classes to novel classes by mistake, which may have disastrous consequences in safety-sensitive applications, such as autonomous driving~\cite{bozhinoski2019safety}. Meanwhile, the static network is constrained to certain scenarios, as it cannot update itself to adapt to new environments. In addition, training from scratch to adapt to new scenes is extremely time-consuming, and the annotations of old classes are sometimes unavailable, due to privacy constrains.

To solve the {\it closed-set} and {\it static} problem, we propose the {\it open-world semantic segmentation} for LIDAR point clouds, which is composed of two tasks: 
1) open-set semantic segmentation (\zsw{OSeg}) to assign the \textit{unknown} label to novel classes as well as to assign the correct labels to old classes, 
and 2) incremental learning (IL) to gradually incorporate the novel classes into the knowledge base after labellers provide the labels of novel classes. Fig.~\ref{fig:example_1} illustrates an example of open-world semantic segmentation for LIDAR point clouds.

As we are the first to study \zsw{OSeg task} 
in the 3D LIDAR point cloud domain, we refer to the existing methods in the 2D image domain, which can be divided into two types, generative network-based methods~\cite{baur2018deep, Lis2019, xia2020synthesize} and uncertainty-based methods~\cite{hendrycks2016baseline, gal2016dropout, DBLP:conf/nips/Lakshminarayanan17}, though none of them can be directly utilized. Generative network-based methods adopt a conditional generative adversarial network (cGAN)~\cite{park2019semantic} to reconstruct the input based on the closed-set prediction results, and assume the novel regions have a larger difference in appearance between the reconstructed input and original input. However, cGAN is not appropriate for reconstruction of the point cloud as all information is determined by the geometry information, \textit{i.e.}, coordinates of points, and cGAN can only reconstruct the channel information, \textit{i.e.}, RGB values, while keeping the geometry information, including coordinates of pixels and the shape of an image, unchanged. The uncertainty-based methods also work poorly as we find the network predicts the novel classes as old classes with high confidence scores, as shown in Fig.~\ref{fig:dis} (a).

In addition to the challenges of the \zsw{OSeg} task, the catastrophic forgetting of old classes in incremental learning~\cite{mccloskey1989catastrophic} is another problem to solve. 
Directly finetuning the network using only the labels of novel classes will make the network classify everything as novel classes. Thus a method is needed to incrementally learn novel classes while keeping the performance of the old classes.


We find that the closed-set and static properties of the traditional closed-set model is due to the fixed classifier architecture, \textit{i.e.}, one classifier corresponds to one old class. 
Therefore, we propose \zsw{a \textbf{RE}dund\textbf{A}ncy c\textbf{L}assifier (REAL) framework} 
to provide a dynamic classifier architecture to adapt the model to both the OSeg and IL tasks. For the OSeg task, we add several redundancy classifiers (RCs) on the basis of the original network to predict the probability of the unknown class. 
Then, during the IL task, several RCs are trained to classify the newly introduced classes, while the remaining RCs are still responsible for the unknown class, as shown in Fig.~\ref{fig:REAL}. 
We provide the training strategies for the OSeg and IL tasks under REAL, based on the unknown object synthesis, predictive distribution calibration, and pseudo label generation. We show the effectiveness of REAL and corresponding training strategies through our comprehensive experiments. 
In summary, our contributions are three-folds:

\begin{itemize}[leftmargin=*, itemsep=0 pt, topsep=0 pt, parsep=0 pt]
    \item We are the first to define the open-world semantic segmentation problem for LIDAR point clouds, which is composed of \zsw{OSeg and IL tasks;}
    \item We propose \zsw{a REAL model} 
    to provide a general architecture for both the OSeg and IL tasks, as well as training strategies for each task, based on the unknown objects synthesis, predictive distribution calibration, and pseudo labels generation;
    \item We construct benchmark and evaluation protocols for OSeg and IL in the 3D LIDAR point cloud domain, based on the SemanticKITTI and nuScenes datasets, to measure the effectiveness of our training strategies under \zsw{REAL}. 
\end{itemize}

\section{Related Work}
\label{sec:related}


\noindent \textbf{Closed-set LIDAR Semantic Segmentation:} Semantic segmentation for LIDAR point clouds can be categorized into point-based and voxel-based methods. Typical point-based methods~\cite{Randla-Net, thomas2019kpconv, wu2019pointconv} use PointNet~\cite{qi2017pointnet} and PointNet++~\cite{qi2017pointnet++} to directly operate on the LIDAR point cloud. However, they have limited performance due to the varying density and large scale of the LIDAR point cloud. The other type of point-based methods convert the LIDAR point cloud to 2D grids and then apply 2D convolutional operations for semantic segmentation. SqueezeSeg~\cite{wu2019squeezesegv2} and RangeNet++~\cite{rangenet++} convert the point cloud to a range image while PolarNet~\cite{polarnet} converts the point cloud to the bird's-eye-view under the polar coordinates. However, 2D representations inevitably lose some of the 3D topology and geometric information. Cylinder3D~\cite{cylindrical} is a voxel-based method and it tackles the sparsity and varying density problems of LIDAR point clouds through cylindrical partition and asymmetrical 3D convolutional networks. Cylinder3D achieves state-of-the-art performance on SemanticKITTI~\cite{se1, se2, se3} and nuScenes~\cite{nuscenes}, so we adopt it as the base architecture in our work.


\noindent \textbf{Open-set 2D Classification:} There are two trends of open-set 2D classification methods: uncertainty-based methods and generative model-based methods. Maximum softmax probability (MSP)~\cite{hendrycks2016baseline} is the baseline of uncertainty-based methods, while Dan \textit{et al.}~\cite{hendrycks2019scaling} found that Maximum Logit (MaxLogit) is a better choice than the probability. MC-Dropout~\cite{gal2016dropout} and Ensembles~\cite{DBLP:conf/nips/Lakshminarayanan17} are used to approximate Bayesian inference~\cite{mackay1995bayesian, DBLP:conf/nips/KendallG17}, which regards the network from a probabilistic view. Meanwhile, generative-based methods, including SynthCP~\cite{xia2020synthesize} and DUIR~\cite{Lis2019}, adopt conditional GAN (cGAN)~\cite{park2019semantic} to reconstruct the input, and find the novel regions by comparing the reconstructed input with the original input. However, these methods cannot adapt to the 3D LIDAR point cloud domain directly, as discussed in Sec.~\ref{sec:intro}. \cite{zhou2021learning, Wang_2021_ICCV} propose to use redundancy classifiers (RCs) to directly output the score of the unknown class, and adopt manifold mixup and a sampler based on Stochastic Gradient Langevin Dynamics (SGLD)~\cite{welling2011bayesian} to approximate the unknown class distribution. We draw inspiration from them, and take a step further by using RCs for both OSeg and IL, as well as developing suitable training strategies for the 3D point cloud domain.


\noindent \textbf{Open-world Classification and Detection:} The open-world problem was first proposed by Abhijit \textit{et al.}~\cite{bendale2015towards}, who argued that the network should be able to deal with a dynamic category set which is practical in the real world. Therefore, they introduced the open-world classification pipeline: first identify both known and unknown images, and then gradually learn to classify unknown images when labels are given. They presented the Nearest Non-Outlier method to manage the open-world classification task. Joseph \textit{et~al.} \cite{joseph2021open} extended the open-world problem to the 2D object detection domain, and proposed a methodology which is based on contrastive clustering, an unknown-aware proposal network and energy-based unknown identification to address the challenges of open-world detection. Jun \textit{et al.}~\cite{Cen_2021_ICCV} later adopted deep metric learning for open-world semantic segmentation for 2D images. Here, we extend the open-world problem to the 3D LIDAR cloud point domain, and both sub-tasks including OSeg and IL for 3D LIDAR point clouds are not studied yet.

\section{Open-world Semantic Segmentation}
\label{sec:ow-def}

In this section, we formalise the definition of open-world semantic segmentation for LIDAR point clouds. Let the classes of the training set be called old classes and labeled by positive integers $\mathcal{K}_0=\left \{ 1,2,...,C \right \} \subset \mathbb{N}^+$. Unlike the traditional closed-set semantic segmentation where the classes of the test set are the same as the training set, some novel classes $\mathcal{U}=\left \{ C+1,... \right \}$ are involved in the test set in the open-world semantic segmentation problem. Let one LIDAR point cloud sample be formulated as $\mathcal{D}=\left \{ \mathbf{P},\mathbf{Y} \right \}$, where $\mathbf{P}=\left \{ \mathbf{p}_1,\mathbf{p}_2,...,\mathbf{p}_M \right \}$ is the input LIDAR point cloud composed of $M$ points and every point $\mathbf{p}$ is represented by three coordinates $\mathbf{p}=(x,y,z)$. The label $\mathbf{Y}=\left \{ y_1,y_2,...,y_M \right \}$ contains the semantic class for every point, in which $y \in \mathcal{K}_0$ for the training data and $y \in \mathcal{K}_0 \cup \mathcal{U}$ for the test data.

Suppose we already have a model $\mathcal{M}_c$ which is trained under the closed-set condition, so its outputs are within the domain of $\mathcal{K}_0$. As discussed in Sec.~\ref{sec:intro}, the open-world semantic segmentation is composed of two tasks: open-set semantic segmentation (OSeg) and incremental learning (IL). For the OSeg task, the model $\mathcal{M}_c$ will be finetuned to $\mathcal{M}_o$ so that it can assign the correct labels for the points of old classes $\mathcal{K}_0$, as well as assign the \textit{unknown} label to the points of novel classes $\mathcal{U}$. For the IL task, the model $\mathcal{M}_o$ will be further finetuned to $\mathcal{M}_i$ when the labels of novel classes $\mathcal{K}_n$ are given, so that its knowledge base is enlarged from $\mathcal{K}_0$ to $\mathcal{K}_0 \cup \mathcal{K}_n$, where $\mathcal{K}_n=\left \{C+1,...,C+n \right \}$. So the classes in $\mathcal{K}_n$ change from \textit{unknown} to \textit{known} for the network. We follow the classical task IL setting~\cite{ilsurvey, peng, cermelli2020modeling} that the new given labels only contain the annotation of the novel class $\mathcal{K}_n$, while the remaining points of old classes $\mathcal{K}_0$ are not annotated. Additionally, the model after IL $\mathcal{M}_i$ still keeps the open-set property, \textit{i.e.}, assigns the \textit{unknown} label to the remaining novel classes $\mathcal{K}_{rn}=\left \{C+n+1,...\right \}$.

\begin{figure}[t]
\centering
\includegraphics[width=0.99\textwidth]{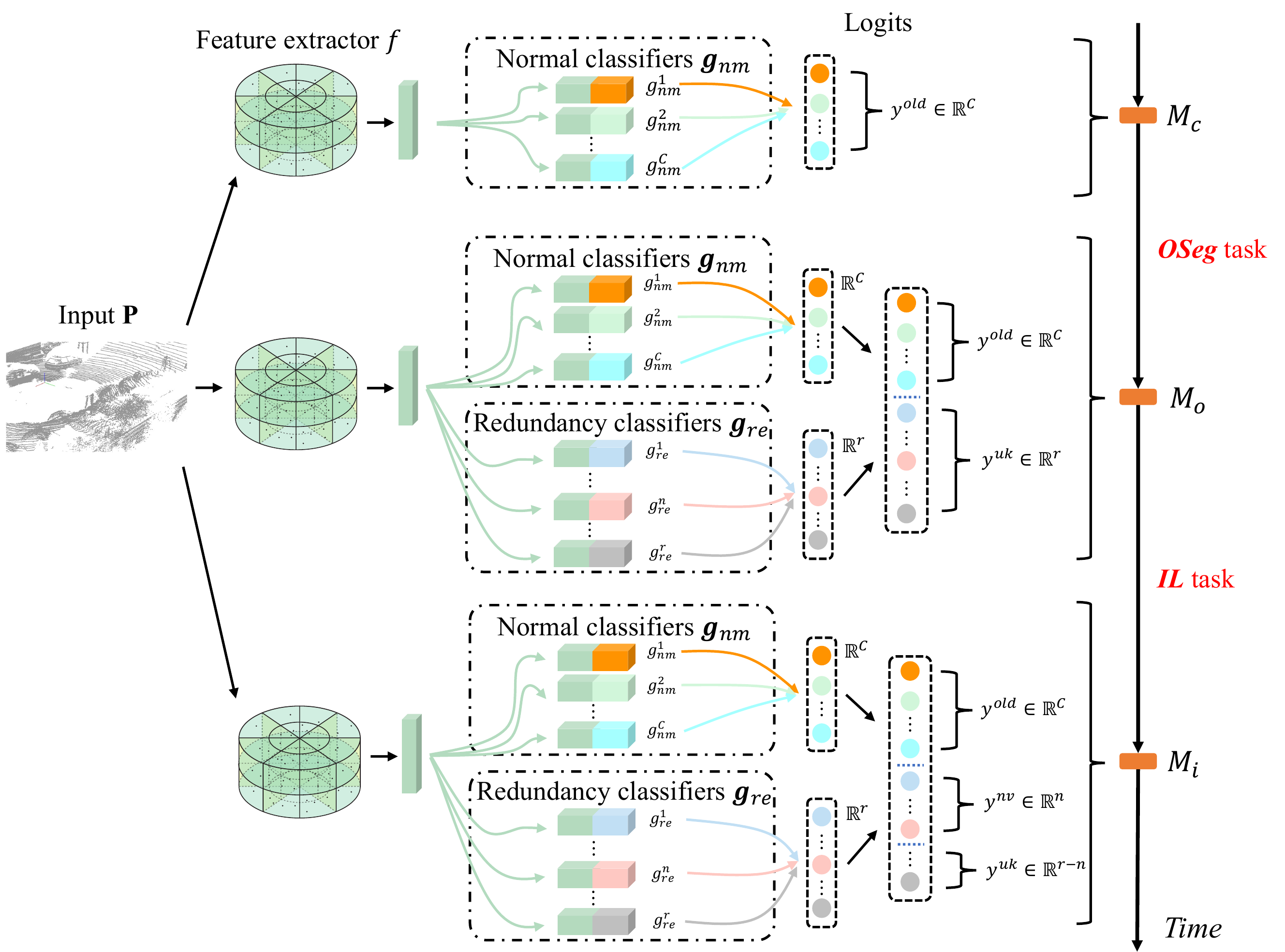}
\vspace{-0.3cm}
\caption{Redundancy classifier framework (REAL). Closed-set model $\mathcal{M}_c$ can only output logits for old classes $y^{old}$. Redundancy Classifiers $g_{re}$ are added on top of the original framework in our REAL. All $g_{re}$ in $\mathcal{M}_o$ are used to output the scores $y^{uk}$ for the unknown class. After the IL task, part of $g_{re}$ are used to output logits for the newly introduced classes $y^{nv}$, while the remaining are still for the unknown class $y^{uk}$.}
\label{fig:REAL}
\end{figure}

\section{Methodology}
In this section, we introduce our strategies to solve the open-world semantic segmentation problem for LIDAR point clouds. The open-world semantic segmentation is composed of two tasks: OSeg task and IL task. We first introduce the redundancy classifier framework (REAL) in Sec.~\ref{sec:REAL}, which provides a general network architecture for both the OSeg task and IL task. Then, we introduce the training strategies and inference procedures for the OSeg task and IL task in Sec.~\ref{sec:open} and Sec.~\ref{sec:incre} respectively.

\subsection{Redundancy Classifier Framework (REAL)}
\label{sec:REAL}

The overall view of REAL is shown in Fig.~\ref{fig:REAL}. The trained closed-set model $\mathcal{M}_c$, which can well classify old classes $\mathcal{K}_0$, is composed of a feature extractor $f$ and normal classifiers $g_{nm}=\left \{ g_{nm}^1, g_{nm}^2,...,g_{nm}^C \right \}$. For a certain input $\mathbf{P} \in \mathbb{R}^{M \times 3}$, the output of the model $\mathcal{M}_c$ is
\begin{equation}
    \mathcal{M}_{c}(\mathbf{P})=[y^{old}]=[g_{nm}(f(\mathbf{P}))] \in \mathbb{R}^{M \times C}.
\end{equation}

\noindent {\bf OSeg task:} The OSeg task is to adapt closed-set model $\mathcal{M}_{c}$ to open-set model $\mathcal{M}_{o}$ so that $\mathcal{M}_{o}$ can identify novel classes $\mathcal{U}$ as {\it unknown}. To achieve this goal, we add $r$ redundancy classifiers (RCs) $g_{re}=\left \{ g_{re}^1, g_{re}^2,...,g_{re}^r \right \}$ on top of the original feature extractor $f$, as shown in Fig.~\ref{fig:REAL} $\mathcal{M}_{o}$. All RCs in $\mathcal{M}_{o}$ are used to predict the scores $y^{uk}$ for the unknown class. We let the maximum response of $y^{uk}$ be the score of the unknown class, which is represented by class $0$. In this way, the output of the open-set model $\mathcal{M}_{o}$ is
\begin{equation}
    \mathcal{M}_{o}(\mathbf{P})=[\max y^{uk}, y^{old}]=[\max \ {g_{re}(f(\mathbf{P}))}, g_{nm}(f(\mathbf{P}))] \in \mathbb{R}^{M \times (1+C)}.
    \label{eq:os_uk}
\end{equation}

\noindent \textbf{IL task:} The IL task is to train open-set model $\mathcal{M}_{o}$ to $\mathcal{M}_{i}$ so that newly introduced classes $\mathcal{K}_n$ change from \textit{unknown} to \textit{known}. $\mathcal{M}_{i}$ is still open-set, {\it i.e.}, it can classify remaining novel classes $\mathcal{K}_{rn}$ as \textit{unknown}. In this task, among all RCs $g_{re}$, some of the RCs $g_{re}^{nv}=\left \{ g_{re}^{1}, g_{re}^{2},...,g_{re}^{n} \right \}$ are used to classify newly introduced classes $\mathcal{K}_n$, {\it i.e.}, $y^{nv}$ in Fig.~\ref{fig:REAL} $\mathcal{M}_{i}$, and the remaining RCs $g_{re}^{uk}=\left \{ g_{re}^{n+1}, g_{re}^{n+2},...,g_{re}^{r} \right \}$ are kept for the unknown class $\mathcal{K}_{rn}$, {\it i.e.}, $y^{uk}$ in Fig.~\ref{fig:REAL} $\mathcal{M}_{i}$. In this way, the output of $\mathcal{M}_{i}$ can be represented as
\begin{equation}
    \mathcal{M}_{i}(\mathbf{P})=[\max y^{uk}, y^{old}, y^{nv}]=[\max \ {g_{re}^{uk}(f(\mathbf{P}))}, g_{nm}(f(\mathbf{P})), g_{re}^{nv}(f(\mathbf{P}))].
    \label{eq:il_uk}
\end{equation}
where $\mathcal{M}_{i}(\mathbf{P}) \in \mathbb{R}^{M \times (1+C+n)}$.

\subsection{Open-set Semantic Segmentation (OSeg)}
\label{sec:open}

The OSeg task is to train the closed-set model $\mathcal{M}_c$ to the open-set model $\mathcal{M}_o$ which can identify novel classes $\mathcal{U}$ as {\it unknown}, as shown in Fig.~\ref{fig:example_1} (c). The network architecture of $\mathcal{M}_o$ is shown in Fig.~\ref{fig:REAL} $\mathcal{M}_o$. We introduce two training methods including {\it Unknown Object Synthesis} and {\it Predictive Distribution Calibration} as well as inference procedure in this section.

\noindent {\bf Unknown Object Synthesis:} We synthesize pseudo unknown objects in the LIDAR point cloud to approximate the distribution of real novel objects. The synthesis process should meet two requirements: 1) the synthesized object should share some invariant basic geometry features with existing objects, such as curved and flat surfaces, so that it can be regarded as an \textit{object} rather than noise and possibly have a similar appearance to real unknown objects; 2) the synthesis process should be as quick as possible.

\begin{figure}[t]
\centering
\includegraphics[width=0.99\textwidth]{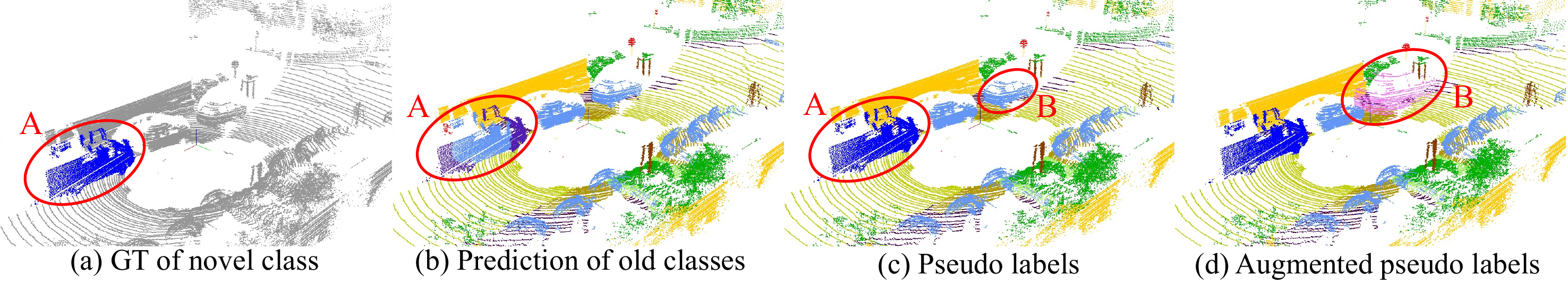}
\vspace{-0.3cm}
\caption{Pseudo labels generating process for incremental learning. Ground truth (a) only contains the label of the novel class (A: other-vehicle). So we combine the prediction results of $\mathcal{M}_o$ (b) to generate the pseudo labels (c). Then we resize objects of old classes as the synthesized objects in (d) (B: resized car).}
\label{fig:method}
\end{figure}

We find that resizing the existing objects with a proper factor is a simple but effective way to conduct the synthesis process, as it keeps the geometric shape of an object, but the different size determines it is a new object. For instance, a car, truck, bus, and construction vehicle have similar local geometric features, such as the shape of the body and tires, but their size can be different. Therefore, we pick up objects of specific old classes $\mathcal{K}_{syn}$ with a probability $p_{syn}$ and resize them from 0.25 to 0.5 times or 1.5 to 3 times as pseudo unknown objects, such as B in Fig.~\ref{fig:method} (c) and (d). In this way, the input $\mathbf{P}$ is divided into two parts: $\mathbf{P}=\mathbf{P}_{syn} \cup \mathbf{P}_{nm}$, where $\mathbf{P}_{syn}$ and $\mathbf{P}_{nm}$ represent the points of synthesized objects and unchanged normal objects respectively. For the points of synthesized objects $\mathbf{P}_{syn}$, the synthesis loss $\mathcal{L}_{syn}$ is
\begin{equation}
    \mathcal{L}_{syn}=\ell (\mathcal{M}(\mathbf{P}_{syn}),\mathbf{0}),
    \label{eq:syn}
\end{equation}
where $\ell$ is the cross-entropy loss. The ground truth labels of synthesized objects are set to be the unknown class $0$, so the first term in Eq.~\ref{eq:os_uk} is trained to give high scores to objects never seen before.

\noindent {\bf Predictive Distribution Calibration:} We find that in the closed-set prediction, the novel objects are classified as old classes with high probability, as shown in Fig.~\ref{fig:dis} (a). We intend to alleviate this problem by probability calibration, and the calibrated scores of the unknown class are shown as Fig.~\ref{fig:dis} (b).
\begin{floatingfigure}[r]{0.5\textwidth}
\centering
\includegraphics[width=0.5\textwidth]{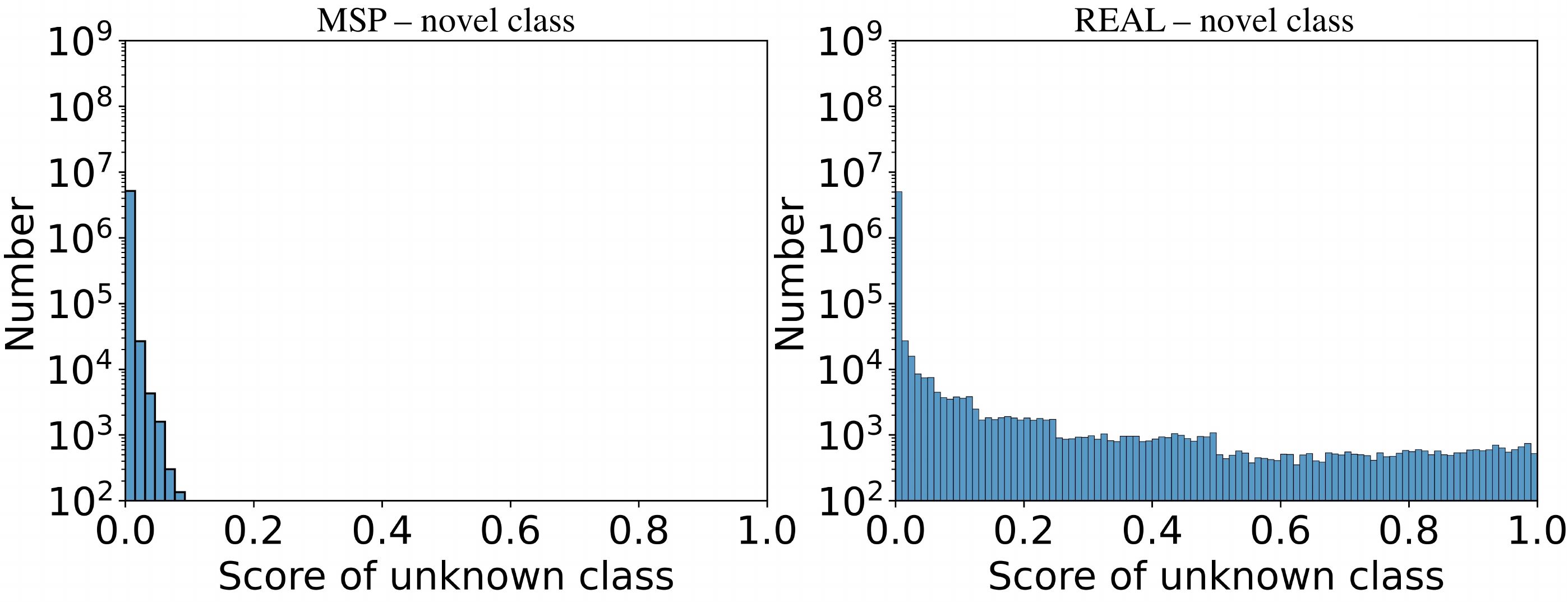}
\caption{Distribution of scores of the unknown class for Maximum Softmax Probability (MSP) and our REAL method. The scores of the unknown class for novel classes are low in MSP (a), meaning the closed-set prediction classifies novel classes as old classes with high confidence.}
\label{fig:dis}
\end{floatingfigure}
We force every point of old classes to have the largest score on its original class, and have the second largest score on the unknown class~\cite{Wang_2021_ICCV}. By this design, the network is supposed to output high probability scores on the unknown class for the novel objects as they do not belong to any one of the old classes. Therefore, for the points of unchanged normal objects $\mathbf{P}_{nm}$, the calibration loss is designed as
\begin{equation}
    \mathcal{L}_{cal}=\mathcal{L}_{cal}^{ori}+\lambda_{cal}\mathcal{L}_{cal}^{uk},
    \label{eq:cal_loss}
\end{equation}
where $\mathcal{L}_{cal}^{ori}$ and $\mathcal{L}_{cal}^{uk}$ are defined as
\begin{equation}
    \mathcal{L}_{cal}^{ori}=\ell (\mathcal{M}(\mathbf{P}_{nm}),\mathbf{Y}_{nm}),
    \label{eq:cal_loss_ori}
\end{equation}
\begin{equation}
    \mathcal{L}_{cal}^{uk}=\ell(\mathcal{M}(\mathbf{P}_{nm}) \ \backslash \ \mathbf{Y}_{nm},\mathbf{0}),
    \label{eq:cal_loss_uk}
\end{equation}
where $\mathbf{Y}_{nm}$ is the ground truth of $\mathbf{P}_{nm}$. $\mathcal{M}(\mathbf{P}_{nm}) \ \backslash \ \mathbf{Y}_{nm}$ means to remove the response of the corresponding ground truth old class. $\mathcal{L}_{cal}^{ori}$ is to ensure the good closed-set prediction, while $\mathcal{L}_{cal}^{uk}$ is to make every point have the second largest probability on the unknown class.

\noindent {\bf Loss Function:} The overall loss function to train the model $\mathcal{M}_c$ to $\mathcal{M}_o$ is
\begin{equation}
    \mathcal{L}^{OSeg}=\mathcal{L}_{cal}^{OSeg}+\lambda_{syn}\mathcal{L}_{syn}^{OSeg},
    \label{eq:os-all}
\end{equation}
where $\mathcal{L}_{cal}^{OSeg}$ is determined by Eq.~\ref{eq:cal_loss}, Eq.~\ref{eq:cal_loss_ori}, and Eq.~\ref{eq:cal_loss_uk}, while $\mathcal{L}_{syn}^{OSeg}$ is determined by Eq.~\ref{eq:syn}. All $\mathcal{M}$ in the related terms are $\mathcal{M}_o$ in the OSeg task.

\noindent {\bf Inference:} Both the closed-set and open-set performance of the finetuned model $\mathcal{M}_o$ will be evaluated. For the closed-set prediction, the inference result $\hat{\mathbf{Y}}_{close}$ is defined as
\begin{equation}
    \hat{\mathbf{Y}}_{close} = \mathop{\arg\max}\limits_{i=1,2,...,C} \ g_{nm}(f(\mathbf{P})).
\end{equation}

For the open-set prediction, we have to classify both old classes and the novel class, so the inference result $\hat{\mathbf{Y}}_{open}$ is defined as:
\begin{equation}
    \hat{\mathbf{Y}}_{open} = \left\{
        \begin{array}{cl}
        \mathop{\arg\max}\limits_{i=1,2,...,C} \ g_{nm}(f(\mathbf{P})) & {\lambda_{conf} < \lambda_{th}} \\
        0 & {otherwise,}
        \end{array} \right.
        \label{eq:if_op}
\end{equation}
where $\lambda_{conf}=\max \ {g_{re}(f(\mathbf{P}))}$ is the confidence score of the unknown class, and $\lambda_{th}$ is the threshold. The unknown class is represented by class 0.

\subsection{Incremental Learning (IL)}
\label{sec:incre}

The IL task is to train $\mathcal{M}_o$ to $\mathcal{M}_i$ when the labels of novel classes $\mathcal{K}_n$ are available. $\mathcal{M}_i$ can classify both newly introduced classes $\mathcal{K}_n$ and old classes $\mathcal{K}_0$, as well as identify remaining novel classes $\mathcal{K}_{rn}$ as {\it unknown}. The inference example is shown in Fig.~\ref{fig:example_1} (d) and the architecture is shown in Fig.~\ref{fig:REAL} $\mathcal{M}_i$.

As mentioned in Sec.~\ref{sec:ow-def}, only the labels of introduced novel classes $\mathcal{K}_n$ are given in this task. Therefore, we divide the unchanged normal points $\mathbf{P}_{nm}$ into two parts, $\mathbf{P}_{nm}^{old}$, which belongs to old classes $\mathcal{K}_0$, and $\mathbf{P}_{nm}^{nv}$, which belongs to newly introduced classes $\mathcal{K}_n$, so that $\mathbf{P}_{nm}=\mathbf{P}_{nm}^{old} \cup \mathbf{P}_{nm}^{nv}$. The labels of points $\mathbf{P}_{nm}^{nv}$ are given as $\mathbf{Y}_{nm}^{nv}$, \textit{e.g.}, labels of A in Fig.~\ref{fig:method} (a), but labels of $\mathbf{P}_{nm}^{old}$ are not given, \textit{e.g.}, gray points in Fig.~\ref{fig:method} (a). If we only use $\mathbf{Y}_{nm}^{nv}$ to directly finetune the model, it will classify all points as the newly introduced class as there is only one kind of class in the training process. This is called the catastrophic forgetting and we use {\it Pseudo Label Generation} to solve this problem.

\noindent {\bf Pseudo Label Generation:} We use model $\mathcal{M}_o$ to predict the pseudo labels $\mathbf{pY}_{nm}^{old}$ for $\mathbf{P}_{nm}^{old}$~\cite{cermelli2020modeling, Cen_2021_ICCV}, as shown in Fig.~\ref{fig:method} (b). In this way, the learned knowledge of old classes is preserved in $\mathbf{pY}_{nm}^{old}$ to alleviate the catastrophic forgetting problem. Then we combine $\mathbf{pY}_{nm}^{old}$ with $\mathbf{Y}_{nm}^{nv}$ to generate the pseudo labels of the whole point cloud $\mathbf{Y}_{nm}$, such as in Fig.~\ref{fig:method} (c).

\noindent \textbf{Loss Function:} Note that we keep the open-set property after IL, so the methods in OSeg task including {\it Unknown Object Synthesis} and {\it Predictive Distribution Calibration} are still used in IL task. The overall loss function to train the model $\mathcal{M}_o$ from $\mathcal{M}_i$ is
\begin{equation}
    \mathcal{L}^{il}=\mathcal{L}_{cal}^{il}+\lambda_{syn}\mathcal{L}_{syn}^{il},
    \label{eq:il-all}
\end{equation}
where $\mathcal{L}_{cal}^{il}$ and $\mathcal{L}_{syn}^{il}$ are determined by Eq.~\ref{eq:cal_loss}, Eq.~\ref{eq:cal_loss_ori}, Eq.~\ref{eq:cal_loss_uk}, and Eq.~\ref{eq:syn}. All $\mathcal{M}$ in the related terms are $\mathcal{M}_i$. Note that $\mathbf{Y}_{nm}$ in Eq.~\ref{eq:cal_loss_ori} and Eq.~\ref{eq:cal_loss_uk} are generated as
\begin{equation}
    \mathbf{Y}_{nm}=\mathbf{pY}_{nm}^{old} \cup \mathbf{Y}_{nm}^{nv},
    \label{eq:il_label}
\end{equation}
where $\mathbf{Y}_{nm}^{nv}$ is the ground truth label of newly introduced classes $\mathcal{K}_{n}$ and $\mathbf{pY}_{nm}^{old}$ is the pseudo labels of old classes $\mathcal{K}_{0}$ generated by $\mathcal{M}_o$,
\begin{equation}
    \mathbf{pY}_{nm}^{old}=\mathcal{M}_o(\mathbf{P}_{nm}^{old}).
\end{equation}
The $\mathbf{Y}_{nm}$ in Eq.~\ref{eq:il_label} contains both newly introduced classes $K_n$ and old classes $K_0$, so $\mathcal{M}_i$ can learn new classes without forgetting old classes.

\noindent {\bf Inference:} To evaluate the performance of IL, we only calculate the closed-set prediction results. This is because, for incremental learning we care about how well the catastrophic forgetting problem is alleviated and the new classes are learned, while the ability to classify the unknown class is already evaluated by Eq.~\ref{eq:if_op} in OSeg task, although after IL the model $\mathcal{M}_i$ can still classify the unknown class $\mathcal{K}_{rn}$. The closed-set inference result $\hat{\mathbf{Y'}}_{close}$ is defined as
\begin{equation}
    \hat{\mathbf{Y'}}_{close} = \mathop{\arg\max}\limits_{i=1,2,...,C+n} \ [g_{nm}(f(\mathbf{P}),g_{re}^{nv}(f(\mathbf{P}))].
\end{equation}

\section{Experiments}

We conduct experiments for both tasks of the open-world semantic segmentation, including OSeg and IL tasks. We evaluate our proposed method on two large-scale datasets, SemanticKITTI and nuScenes.

\subsection{Open-world Evaluation Protocol}

\noindent \textbf{Data Split:} We set the novel classes of SemanticKITTI $\mathcal{K}_n^{sk}$ and nuScenes $\mathcal{K}_n^{ns}$ as:
\vspace{-0.2cm}
$$\mathcal{K}_n^{sk}=\left \{ \textit{other-vehicle} \right \}$$
$$\mathcal{K}_n^{ns}=\left \{ \textit{barrier, construction-vehicle, traffic-cone, trailer} \right \}$$
All remaining classes are included in the old class set $\mathcal{K}^{sk}_0$ and $\mathcal{K}^{ns}_0$. During training of the closed-set model $\mathcal{M}_c$ and open-set model $\mathcal{M}_o$, we set the labels of novel classes $\mathcal{K}_n^{sk}$ and $\mathcal{K}_n^{ns}$ to be void and ignore them. During incremental learning, we gradually introduce the labels of novel classes $\mathcal{K}_n^{sk}$ and $\mathcal{K}_n^{ns}$ one by one, and set the labels of old classes $\mathcal{K}^{sk}_0$ and $\mathcal{K}^{ns}_0$ to be void.

\noindent \textbf{Evaluation Metrics:} To evaluate the performance of the open-set semantic segmentation model $\mathcal{M}_o$, we consider both the closed-set and open-set segmentation ability. The closed-set ability is measured by mIoU$\boldsymbol{\mathrm{_{close}}}$, while the open-set evaluation is regarded as a binary classification problem between the known class and unknown class, which is measured by area under the ROC curve (AUROC) and area under the precision-recall curve (AUPR) \cite{hendrycks2019scaling}. 

To evaluate the performance of the model $\mathcal{M}_{i}$ after incremental learning, we calculate the performance of the old classes mIoU$\boldsymbol{\mathrm{_{old}}}$ and newly introduced classes mIoU$\boldsymbol{\mathrm{_{novel}}}$ respectively, and also the mIoU of all classes.

\begin{table}[t]
\begin{center}
\caption{Benchmark of open-set semantic segmentation for LIDAR point clouds. Results are evaluated on the validation set.}
\small
\begin{tabular}{l|ccc|ccc}
\toprule [1pt]
Dataset    & \multicolumn{3}{c|}{SemanticKITTI}                         & \multicolumn{3}{c}{nuScenes}                        \\ \midrule
Methods     & AUPR & AUROC & mIoU$\boldsymbol{\mathrm{_{old}}}$ & AUPR & AUROC & mIoU$\boldsymbol{\mathrm{_{old}}}$\\ \midrule
Closed-set  & 0  & 0  &  58.0 
& 0  & 0   &  58.7 \\
Upper bound &  73.6  &  97.1  & 63.5 
&  86.1  &  99.3   & 73.8  \\ \midrule
MSP  & 6.7  & 74.0  &  58.0  
& 4.3  & 76.7  &  58.7  \\
MaxLogit  & 7.6  & 70.5   &  58.0  
& 8.3  & 79.4   &  58.7  \\
MC-Dropout  & 7.4  & 74.7   &  58.0  
& 14.9  & 82.6   &  58.7  \\
REAL       & \textbf{20.8} & \textbf{84.9}  & 57.8  & \textbf{21.2} & \textbf{84.5} & 56.8 \\
\bottomrule [1pt]
\end{tabular}
\end{center}
\label{tab:1}
\end{table}

\subsection{Open-set Semantic Segmentation (OSeg)}

\noindent \textbf{Implementation:} We adopt Cylinder3D as the base network and train the traditional closed-set model $\mathcal{M}_{c}$ following the training settings in \cite{cylindrical} using the labels of old classes $\mathcal{K}^{sk}_0$ and $\mathcal{K}^{ns}_0$. Then we add several redundancy classifiers on top of the $\mathcal{M}_{0}$ and finetune the model $\mathcal{M}_{c}$ to $\mathcal{M}_{o}$ based on the training strategies described in Sec.~\ref{sec:open}. The old classes used to synthesize novel objects $\mathcal{K}_{syn}$ are \textit{car} for SemanticKITTI and \textit{car, bus,} and \textit{truck} for nuScenes. The probability of resizing these objects $p_{syn}$ is set to 0.5. The unknown object synthesis time is 0.5-4 $ms$ based on our experiments, which is sufficiently quick.

\noindent \textbf{Baselines and Upper Bound:} We refer to several methods from the open-set 2D semantic segmentation domain and implement them in our 3D LIDAR points domain as our baselines, including MSP, Maxlogit, and MC-Dropout, as discussed in Sec.~\ref{sec:related}. The upper bound is to use labels of all classes $\mathcal{K}_0 \cup \mathcal{K}_n$ to train the network and regard the softmax probability of the classes $\mathcal{K}_n$ as the confidence score.

\noindent \textbf{Quantitative results:} The quantitative results of OSeg are shown in Tab.~1. The closed-set method does not consider the unknown class at all, so the open-set evaluation metrics are 0. Among all open-set semantic segmentation baselines, our REAL achieves remarkably better results on the open-set evaluation metrics. The closed-set mIoU$\boldsymbol{\mathrm{_{old}}}$ shows that our method does not sacrifice the ability to classify old classes. The upper bound naturally achieves the best performance as it is conducted in a supervised manner, while the information of the unknown class is not provided for other open-set methods.

\begin{figure}[t]
\centering
\includegraphics[width=0.99\textwidth]{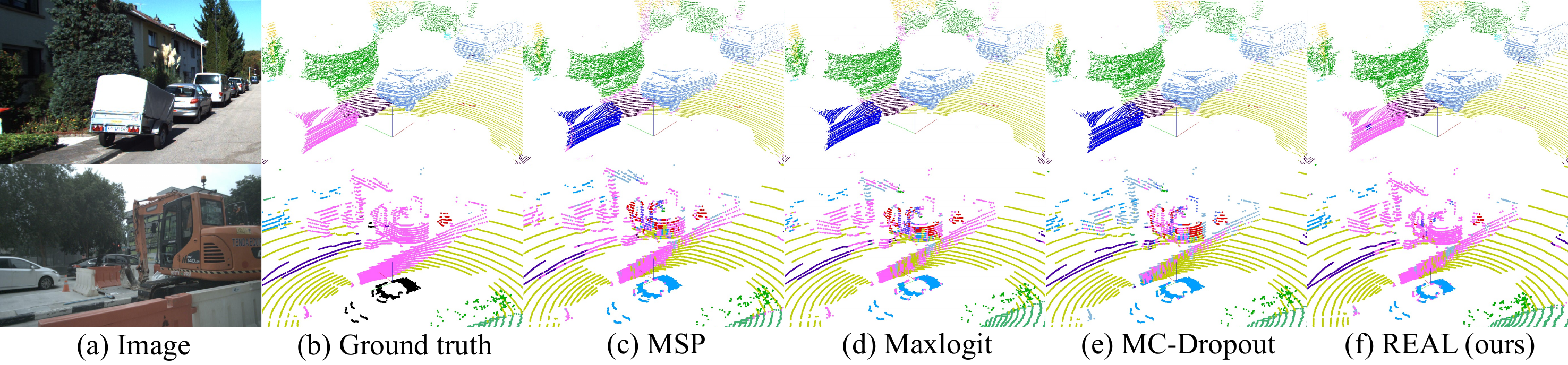}
\vspace{-0.3cm}
\caption{Qualitative results of OSeg task. Novel classes are in pink (other-vehicle in SemanticKITTI (top), and construction-vehicle and barrier in nuScenes (bottom)). The results show that our method has a better performance in distinguishing the novel class from old classes than all the baselines. Best viewed in zoom.}
\label{fig:REAL_ood}
\end{figure}

\begin{figure}[t]
  \centering
  \includegraphics[width=0.99\linewidth]{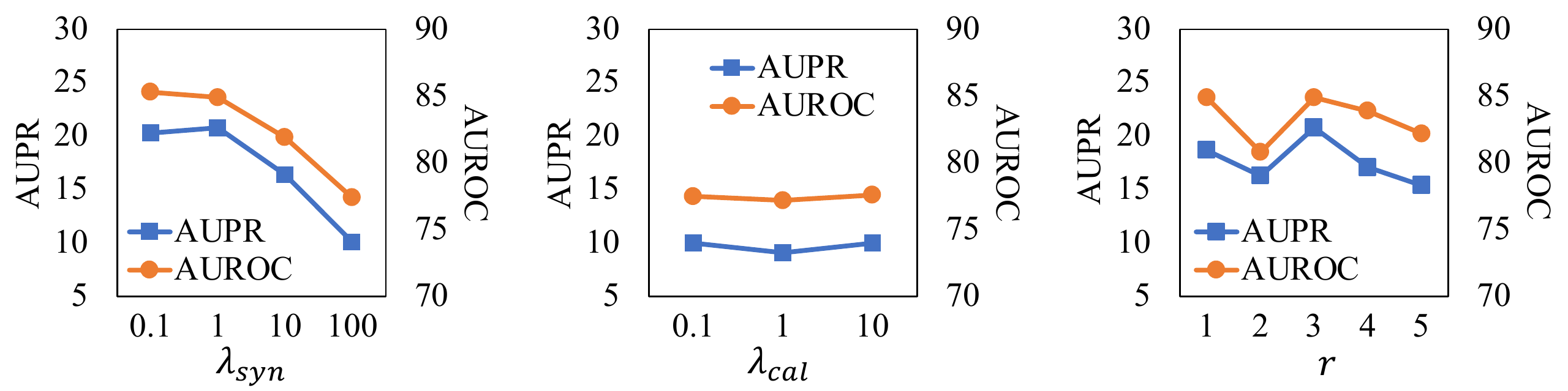}
  \vspace{-0.3cm}
   \caption{Ablation experiments of coefficient $\lambda_{syn}$, $\lambda_{cal}$ and number of redundancy classifiers $r$ for OSeg task on SemanticKITTI.}
   \label{fig:abl}
\end{figure}

\noindent \textbf{Qualitative results:} Fig.~\ref{fig:REAL_ood} contains the qualitative results from SemanticKITTI and nuScenes respectively. Fig.~\ref{fig:REAL_ood} top row shows that our method can identify the other-vehicle as the novel class, while all baselines consider it as the truck. In Fig.~\ref{fig:REAL_ood} bottom row, the baselines classify the construction-vehicle as the truck, pedestrian, and manmade, while our method distinguishes it as the novel object.

\begin{table}[b]
\caption{Ablation study results of $\mathcal{L}_{cal}$ and $\mathcal{L}_{syn}$ for OSeg task on SemanticKITTI.}
\small
\begin{center}
\begin{tabular}{c|cc|ccc}
\toprule [1pt]
Row ID &$\mathcal{L}_{cal}$                  & $\mathcal{L}_{syn}$                  & AUPR & AUROC & mIoU$\boldsymbol{\mathrm{_{old}}}$ \\ \midrule
1&  \small{\xmark}                  & \small{\xmark}       & 0    & 0     & 58.0   \\
2&\checkmark                   & \small{\xmark} & 10.0 & 77.5  & \textbf{58.1} \\
3&\checkmark                     & \checkmark                        & \textbf{20.8} & \textbf{84.9}  & 57.8 \\
\bottomrule [1pt]
\end{tabular}
\end{center}
\label{tab:2}
\end{table}

\noindent \textbf{Ablation experiments:} We carefully conduct ablation experiments on the SemanticKITTI dataset to verify the effectiveness of our we proposed components. According to the results of Row ID 2 in Tab.~\ref{tab:2}, using the calibration loss alone can already outperforms all baselines in Tab.~1. Furthermore, the result of Row ID 3 illustrates that resizing the objects of existing classes with a proper factor is a simple but useful way to imitate novel objects. $\lambda_{syn}$ and $r$ are set to be 1 and 3 according to Fig.~\ref{fig:abl}. $\lambda_{cal}$ is 0.1, and it does not influence the result with a large margin based on Fig.~\ref{fig:abl}.

\subsection{Incremental Learning}
\label{ex_il}

\begin{table}[t]
\caption{\footnotesize Incremental learning results on SemanticKITTI 18 + 1 (other-vehicle) setting.}
\vspace{-0.1cm}
\centering
\begin{tabular}{l|ccc|ccc}
\toprule [1pt]
SemanticKITTI 18+1 & \multicolumn{3}{c|}{Validation set}        & \multicolumn{3}{c}{Test set} \\ \midrule
Method             & mIoU & mIoU$\boldsymbol{\mathrm{_{novel}}}$ & mIoU$\boldsymbol{\mathrm{_{old}}}$ & mIoU & mIoU$\boldsymbol{\mathrm{_{novel}}}$ & mIoU$\boldsymbol{\mathrm{_{old}}}$ \\ \midrule
Closed-set         & 58.0 & 0           & 61.2      & 61.8       & 0       &    65.3    \\
Upper bound        & 63.5 & 44.1        & 64.6      & 62.2       & 40.1       & 63.5       \\ \midrule
Finetune           & 0    & 0.5           & 0         &  0      &     0   & 0       \\
Feature extraction & 6.8 & 0.6         & 7.1      &  6.9      & 0.4      & 7.3      \\
LwF          & 21.6  & 1.7         & 22.7      &  20.2      & 0.9    & 21.3       \\
REAL                & \textbf{64.3} & \textbf{51.5}   & \textbf{65.0}  & \textbf{61.1}       & \textbf{25.3}    & \textbf{63.1}         \\ \bottomrule [1pt]
\end{tabular}
\label{tab:3}
\end{table}

\noindent \textbf{Implementation:} We adopt the training strategies described in Sec.~\ref{sec:incre} to finetune the model $\mathcal{M}_{o}$ to $\mathcal{M}_{i}$. The old classes used for synthesis $\mathcal{K}_{syn}$ are the same as the set during training from $\mathcal{M}_{c}$ to $\mathcal{M}_{o}$.

\begin{table}[b]
\caption{\footnotesize Incremental learning results on nuScenes for 12 + 4 (barrier, construction-vehicle, traffic-cone, and trailer) setting.}
\centering
\begin{tabular}{l|ccc|ccc}
\toprule [1pt]
nuScenes 12+4 & \multicolumn{3}{c|}{Validation set}        & \multicolumn{3}{c}{Test set} \\ \midrule
Method             & mIoU & mIoU$\boldsymbol{\mathrm{_{novel}}}$ & mIoU$\boldsymbol{\mathrm{_{old}}}$ & mIoU & mIoU$\boldsymbol{\mathrm{_{novel}}}$ & mIoU$\boldsymbol{\mathrm{_{old}}}$ \\ \midrule
Closed-set         & 58.7 & 0           & 78.3      & 55.8       & 0       &    74.4    \\
Upper bound        & 73.8 & 62.5        & 77.6      & 73.8       & 70.4       & 74.8       \\ \midrule
Finetune           & 0    & 0           & 0         &  0      &     0   & 0       \\
Feature extraction & 5.5 & 2.1         & 6.6      &  5.3      & 1.9      & 6.4      \\
LwF          & 6.1  & 2.4         & 7.3      &  5.6      & 2.5    & 6.6       \\
REAL                & \textbf{74.9} & \textbf{62.2}   & \textbf{79.1}  & \textbf{74.2}       & \textbf{71.9}    & \textbf{75.0}         \\ \bottomrule [1pt]
\end{tabular}
\label{tab:4}
\end{table}

\noindent \textbf{Baselines and upper bound:} We adopt direct finetuning of $\mathcal{M}_{o}$ to $\mathcal{M}_{i}$ using only the labels of novel classes $\mathcal{K}_{n}^{sk}$ and $\mathcal{K}_{n}^{ns}$ to illustrate the catastrophic forgetting problem. Two methods including Feature Extraction and Learning without Forgetting (LwF)~\cite{lwf} using $\mathcal{K}_{n}^{sk}$ and $\mathcal{K}_{n}^{ns}$ are regarded as the baselines. The upper bound is the same as the upper bound in the open-set semantic segmentation task, which uses all labels $\mathcal{K}_0 \cup \mathcal{K}_n$ to train the network.

\noindent \textbf{Quantitative results:} Tab.~\ref{tab:3} and Tab.~\ref{tab:4} show the IL performance of SemanticKITTI and nuScenes dataset respectively. Directly finetuning the model $\mathcal{M}_{o}$ to $\mathcal{M}_{i}$ only using labels of the novel class incurs the catastrophic forgetting problem, \textit{i.e.}, the network classifies all points as the new class. mIoU$\boldsymbol{\mathrm{_{old}}}$ becomes 0 as there is no prediction results in old classes. mIoU$\boldsymbol{\mathrm{_{novel}}}$ is also close to 0 as newly introduced class only counts a little portion in the whole point cloud. In contrast, mIoU$\boldsymbol{\mathrm{_{old}}}$ in our method is similar with the closed-set, meaning our method can learn the new classes one by one without forgetting the old classes. Our methods has better performance compared to two baselines, showing that using the unlabeled background points $\mathbf{Y}_{nm}^{old}$ is extremely helpful to preserve the old knowledge. Compared to the upper bound, our method only needs the ground truth of newly introduced classes $\mathcal{K}_n$ and consumes much less time in training (5 epochs v.s. 35 epochs), while keeping the similar performance.

\begin{figure}[t]
  \centering
  \includegraphics[width=0.99\linewidth]{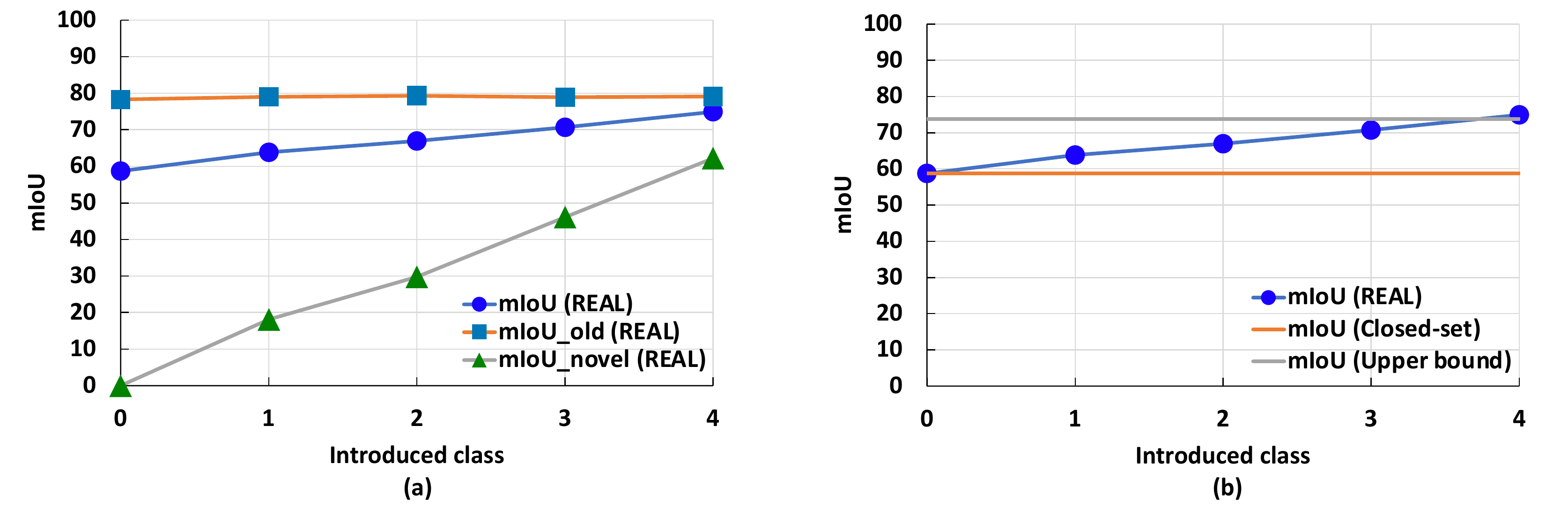}
  \vspace{-0.3cm}
   \caption{Incremental learning results for nuScenes validation set. Introduced class: 1: barrier; 2: construction-vehicle; 3: traffic-cone; 4: trailer.}
   \label{fig:supp}
\end{figure}

\begin{table}[b]
\caption{Incremental learning results on nuScenes test set for 12 + 4 setting. $\text{REAL}_{1}$ to $\text{REAL}_{4}$ means we introduce the label of one novel class per time and conduct incremental learning one by one.}
\centering
\scriptsize
\renewcommand\arraystretch{1}
\renewcommand\tabcolsep{0.5pt}
\begin{tabular}{l|cccccccccccc|cccc|ccc}
\toprule [1pt]
nuScenes 12+4 & \multicolumn{12}{c|}{Old classes}        & \multicolumn{4}{c|}{Novel classes} & \multicolumn{3}{c}{Metrics} \\ \midrule
 Method & \rotatebox{90}{bicycle} & \rotatebox{90}{bus} & \rotatebox{90}{car} &  \rotatebox{90}{motorcycle} & \rotatebox{90}{pedestrian} & \rotatebox{90}{truck}  & \rotatebox{90}{driveable} & \rotatebox{90}{other-flat} & \rotatebox{90}{sidewalk} & \rotatebox{90}{terrain} & \rotatebox{90}{manmade} & \rotatebox{90}{vegetation} &  \rotatebox{90}{barrier} & \rotatebox{90}{construction} & \rotatebox{90}{traffic-cone} & \rotatebox{90}{trailer} & \rotatebox{90}{mIoU} & \rotatebox{90}{mIoU$\boldsymbol{\mathrm{_{novel}}}$} & \rotatebox{90}{mIoU$\boldsymbol{\mathrm{_{old}}}$} \\ \midrule
Upper bound                                      & 26.7 & 83.7     & 84.7     & 72.6  & 73.7          & 68.8       & 96.9    & 68.8 & 75.7   & 71.0  & 88.3  & 86.4      & 80.9    & 55.1       & 67.4 & 79.4  &73.8        &70.7       & 74.8                               \\ \midrule
Closed-set                                       & 28.0 & 83.0     & 86.3     & 75.8 & 74.1  & 58.6 & 96.9          & 67.2         & 77.1       & 72.6    & 86.6 & 86.0   & 0.0  & 0.0  & 0.0      &0.0    & 55.8    & 0 & 74.4                      \\
 $\text{REAL}_{1}$                                & 28.1 & 83.6     & 86.7     & 78.1   & 75.0  & 58.4 & 97.1           & 67.3         & 77.6       & 74.0     & 88.3 & 87.2     & 80.9 &0.0 &0.0 &0.0  & 61.4    & 20.2     & 75.1 \\
$\text{REAL}_{2}$                                & 27.3 & 82.5     & 86.3     & 77.7   & 75.4  & 58.6 & 97.1           & 67.0         & 77.4       & 73.3     & 88.1 & 86.8     & 80.9 &56.7 &0.0 &0.0  & 64.7    & 34.4     & 74.8 \\
$\text{REAL}_{3}$                                & 28.1 & 82.1     & 86.1     & 77.7   & 75.4  & 57.6 & 97.0           & 66.3         & 77.4       & 73.7     & 88.4 & 87.0     & 81.1 &57.6 &66.5 &0.0  & 68.9    & 51.3     & 74.8 \\
$\text{REAL}_{4}$                                & 23.3 & 82.7     & 85.8     & 76.2   & 75.0  & 68.8 & 96.9           & 65.0         & 77.4       & 73.2     & 88.7 & 86.8     & 81.2 &58.6 &66.8 &81.1  & 74.2    & 71.9     & 75.0 \\
\bottomrule [1pt]
\end{tabular}
\label{tab:5}
\end{table}

We show the performance of the model on the nuScenes dataset during IL in Fig.~\ref{fig:supp} and Tab.~\ref{tab:5}. Fig.~\ref{fig:supp} (a) shows during IL the model are gradually learning novel classes while keeping the performance of old classes. Fig.~\ref{fig:supp} (b) illustrates the model starts from the closed-set model and finally achieves the comparable performance with the upper bound. 

\subsection{Open-world Semantic Segmentation}

\begin{figure}[t]
  \centering
  \includegraphics[width=0.99\linewidth]{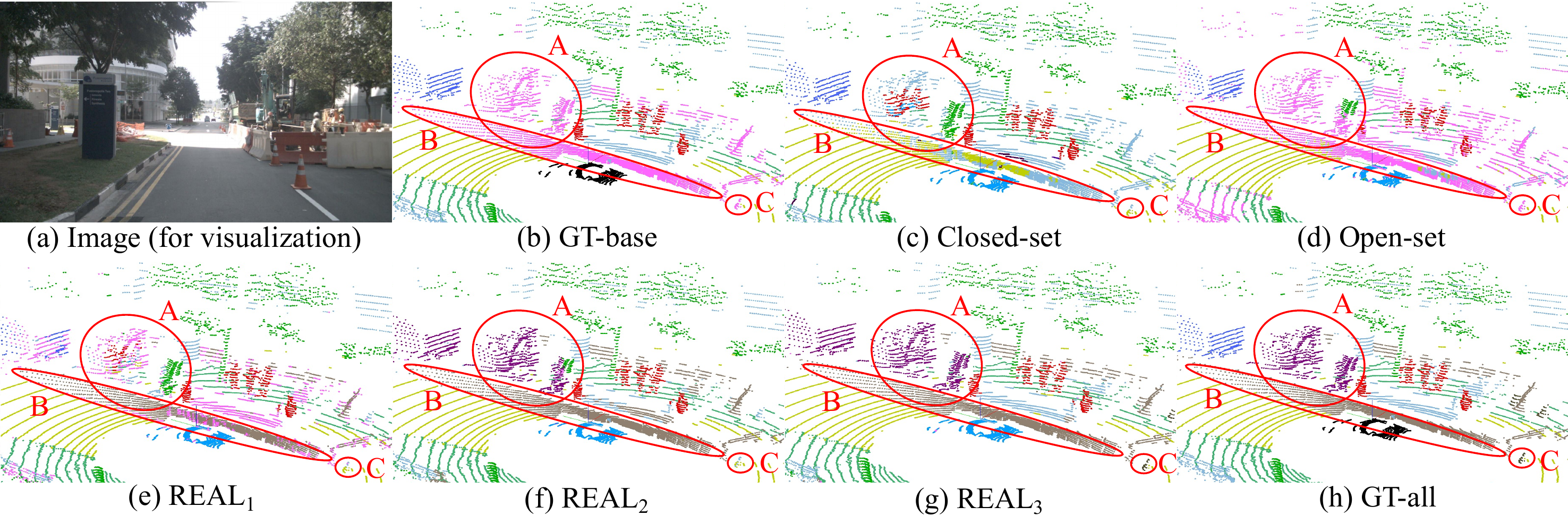}
  \vspace{-0.3cm}
   \caption{Qualitative results of open-world semantic segmentation. GT: ground truth. In (b) GT-base we set the novel classes $\mathcal{K}_n$ in pink (\textbf{A}: construction-vehicle; \textbf{B}: barrier; \textbf{C}: traffic-cone). (c) Closed-set prediction classifies novel objects as old classes. (d) Open-set prediction can identify these novel objects as {\it unknown}. We gradually introduce the labels of barrier, construction-vehicle, and traffic-cone in (e) $\text{REAL}_1$, (f) $\text{REAL}_2$, and (g) $\text{REAL}_3$, so they can classify these novel classes one by one. (h) GT-all contains ground truth of all classes.}
   \label{fig:incre}
\end{figure}

We illustrate the whole open-world semantic segmentation system in Fig.~\ref{fig:incre}. Traditional closed-set model $\mathcal{M}_c$ classifies objects of novel classes $\mathcal{K}_n$ as old classes $\mathcal{K}_0$. In Fig.~\ref{fig:incre} (c), A (construction vehicle) is classified as manmade, pedestrian, and truck; B (barrier) is classified as road and manmade; C (traffic-cone) is classified as road. Such misclassification may cause serious problems in autonomous driving. Thus we conduct the methods in Eq.~\ref{eq:os-all} to finetune $\mathcal{M}_c$ to $\mathcal{M}_o$ so that this open-set model can identify these novel objects as {\it unknown}, as shown in pink area of Fig.~\ref{fig:incre} (d). Then, after incremental learning using the methods described in Eq.~\ref{eq:il-all}, the model can gradually classify new classes, \textit{e.g.}, A (barrier), B (construction-vehicle), and C (traffic-cone) in Fig.~\ref{fig:incre} (e), (f), and (g). Note that after incremental learning the model can still identify unknown classes, as shown in the pink areas of Fig.~\ref{fig:incre} (e).

\section{Conclusion}

Traditional closed-set semantic segmentation cannot handle objects of novel classes. In this paper, we propose the open-world semantic segmentation for LIDAR point clouds, where the model can identify novel objects (open-set semantic segmentation) and then gradually learn them when labels are available (incremental learning). We propose the redundancy classifier framework (REAL) and corresponding training and inference strategies to fulfill the open-world semantic segmentation system. We hope this work can draw the attention of researchers toward this meaningful and open problem.

\clearpage
%
%
\bibliographystyle{splncs04}
\bibliography{egbib}
\end{document}